\let\svthefootnote\thefootnote
\newcommand\freefootnote[1]{%
  \let\thefootnote\relax%
  \footnotetext{#1}%
  \let\thefootnote\svthefootnote%
}

\documentclass[cameraready]{Interspeech}

\title{WARDEN: Endangered Indigenous Language Transcription and Translation with 6 Hours of Training Data}

\author{Ziheng}{Zhang$^*$$^a$}
\author{Yunzhong}{Hou$^*$$^a$}
\author{Naijing}{Liu$^b$}
\author{Liang}{Zheng$^a$}

\address{
    $^a$ Australian National University
    $^b$ University of Oxford
}

\email{ziheng.zhang@anu.edu.au, hou\_yz@outlook.com, naijing.liu@ling-phil.ox.ac.uk, liang.zheng@anu.edu.au}

\keywords{speech recognition, large language models, computational paralinguistics}

\usepackage{comment}
\usepackage{pifont}
\usepackage{graphicx}
\usepackage{hyperref}

\begin{document}

\maketitle

\begin{abstract}
\freefootnote{* equal contribution}
This paper introduces WARDEN\footnote{It is abbreviation for Wardaman Decoding ENgine.}, an early language model system capable of transcribing and translating Wardaman, an endangered Australian indigenous language into English. The significant challenge we face is the lack of large-scale training data: in fact, we only have 6 hours of annotated audio. Therefore, while it is common practice to train a single model for transcription and translation using large datasets (like English to French), this practice is no longer viable in the Wardaman to English context. To tackle the low-resource challenge, we design WARDEN to have separate transcription and translation models: WARDEN first turns a Wardaman audio input into phonemic transcription, and then the transcription into English translation. Further, we propose two useful techniques to enhance performance. For transcription, we initialize the Wardaman token from Sundanese, a language that shares similar phonemes with Wardaman, to accelerate fine-tuning of the transcription model. For translation, we compile a Wardaman-English dictionary from expert annotations, and provide this domain-specific knowledge to a large language model (LLM) to reason and decide the final output. We empirically demonstrate that this two-stage design works better than data-hungry unified approaches in extremely low data settings. Using a mere 6 hours of annotated data, WARDEN outperforms larger open-source and proprietary models and establishes a strong baseline. Data and code are available at \href{https://github.com/Ziheng-Zhang-AUS/WARDEN}{\textcolor{blue}{link}}.
\end{abstract}

\section{Introduction}
The world has numerous small languages. For example, the Wardaman language examined in this work is a highly endangered non-Pama-Nyungan language spoken in the Northern Territory of Australia, with only two full speakers as of 2025 \cite{merlan1994narratives,merlan_wardaman_elar}. 

Documenting such languages must be carried out in person by well-trained linguists. An important task in the documentation process is producing time-aligned transcriptions and translations, which relies heavily on collaboration between linguists and local community members. It is well established in language documentation that transcription and translation require a considerable time commitment. For example, it can take days to accurately transcribe and translate just one hour of audio \cite{himmelmann2006language}. Although there were originally only a few ELAN-annotated hours, the ELDP project has since increased that number considerably. As a result, the largest dataset to which we have access still contains only 6 hours of time-aligned transcribed and translated segments, covering more than 10 hours of audiovisual recordings \cite{merlan_wardaman_elar}.



\begin{figure}[t]
    \centering
    \includegraphics[width=\linewidth]{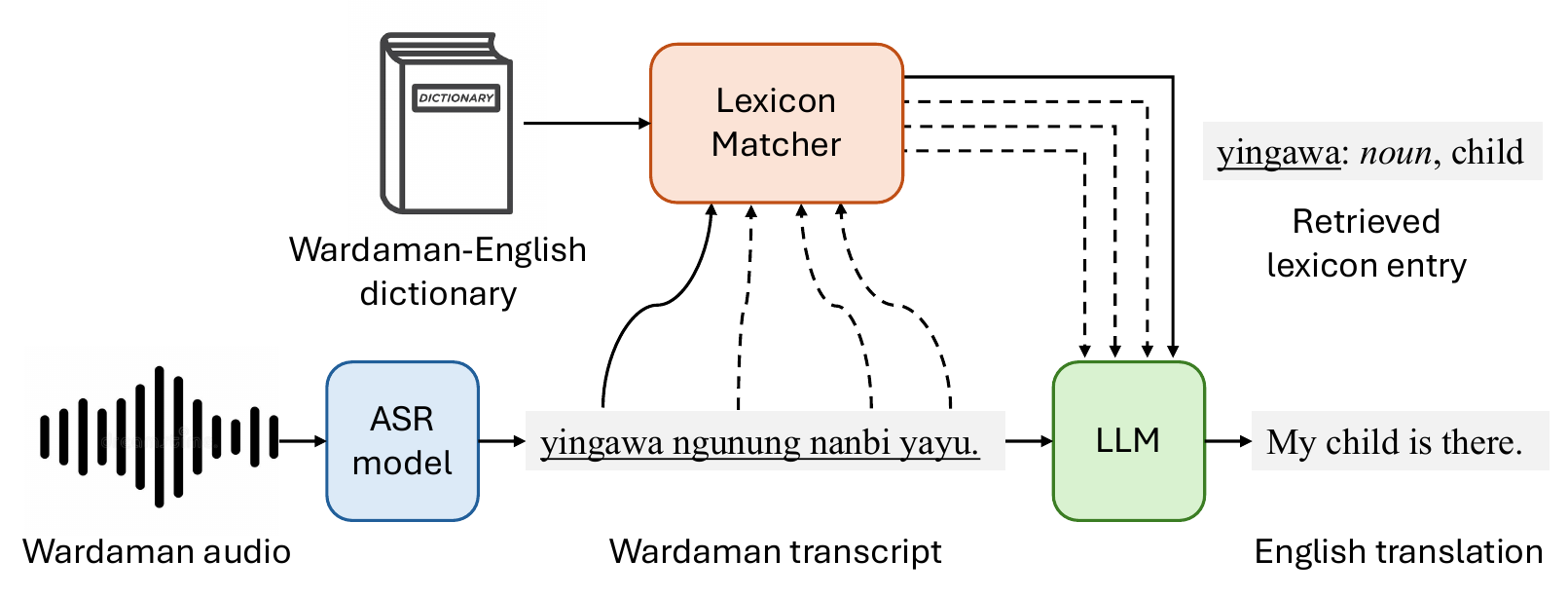}
    \caption{Overview of the WARDEN system.
    For transcription, we select the language most similar to Wardaman for token initialization and fine-tune an existing ASR model. For translation, given transcription results, a lexicon matcher first retrieves relevant Wardaman-English dictionary entries. Then, both the transcript and matched lexicons are fed into an LLM for translation.}
    \label{fig:pipeline}
\end{figure}

A potential way to accelerate this process is to use automatic speech recognition (ASR) \cite{adams2019massively} and machine translation (MT) \cite{bird2022machine}, which can transcribe and translate a large volume of audio data at a relatively low cost. 
However, these automatic systems need large amounts of annotated data for training and thus are built for major languages like English and Chinese. This requirement of training data is infeasible for small languages like Wardaman. 

To automate language documentation of low-resource languages like Wardaman, we introduce WARDEN, a two-stage system that first transcribes and then translates the Wardaman language into English. 
WARDEN adopts two key strategies grounded in linguistic insights. 
\textbf{First}, for transcription, we leverage cross-linguistic phonological similarity and select a proxy language, Sundanese, for language token initialization in the ASR model. This helps to provide a suitable inductive bias and accelerate knowledge transfer during ASR model fine-tuning \cite{conneau2020unsupervised}. 
\textbf{Second}, for translation, we leverage knowledge from an expert-compiled Wardaman-English distionary when fine-tuning a large language model (LLM). 
The LLM takes as input both transcription results from the ASR model and  expert-annotated lexicon entries most related to the transcription results. The LLM is expected to use its general knowledge and reasoning capability to understand the transcript using field-specific knowledge from the lexicons. This approach transforms the LLM from a data-hungry translator into a knowledge-grounded interpreter. The two strategies allow WARDEN to escape the curse of limited data that haunts existing systems \cite{liu2024whisper}. 


Integrating these two components, we introduce WARDEN, 
a framework that separates the transcription and translation for data-efficient learning. 
Trained on only 6 hours of data, WARDEN outperforms data-hungry transcription and translation systems \cite{liu2024whisper}, verifying the effectiveness of our design.

\section{Related Work}
Recent studies have shown that fine-tuning models like Whisper for translation and transcription require a significant amount of data even when dealing with low-resource languages. Liu et al.~\cite{liu2024whisper} report that it requires more than tens of hours of data for Whisper to effectively reduce the word error rate (WER) across seven languages. Timmel \textit{et al.}~\cite{sicard2023swissgerman} show that it takes more than 900 hours of training data to reduce the WER for Swiss German from 45\% to 18\% through fine-tuning, demonstrating the data-hungry nature.
In translation, large language models like Qwen3~\cite{yang2025qwen3}, which often possess strong zero-shot capabilities due to the vast semantic knowledge learned during training, often encounter problems when vertical domain data is scarce and the distribution differs from the pre-training data. To bridge this gap, LexC-Gen~\cite{yong2024lexcgen} attempts to guide the generation of synthetic data using bilingual dictionaries, while Zheng and Yu~\cite{zheng2025tangut} improve BLEU scores in Tangut--Chinese translation with external word-for-word translation.

In contrast to the training data requirements in the scale of tens or hundreds in existing works, we are working with a mere 6 hours of audio data. Thus, we separate the translation and the transcription stages, and introduce inductive bias from expert knowledge to help with model fine-tuning. 

\begin{figure}[t]
  \centering
  \includegraphics[width=\linewidth]{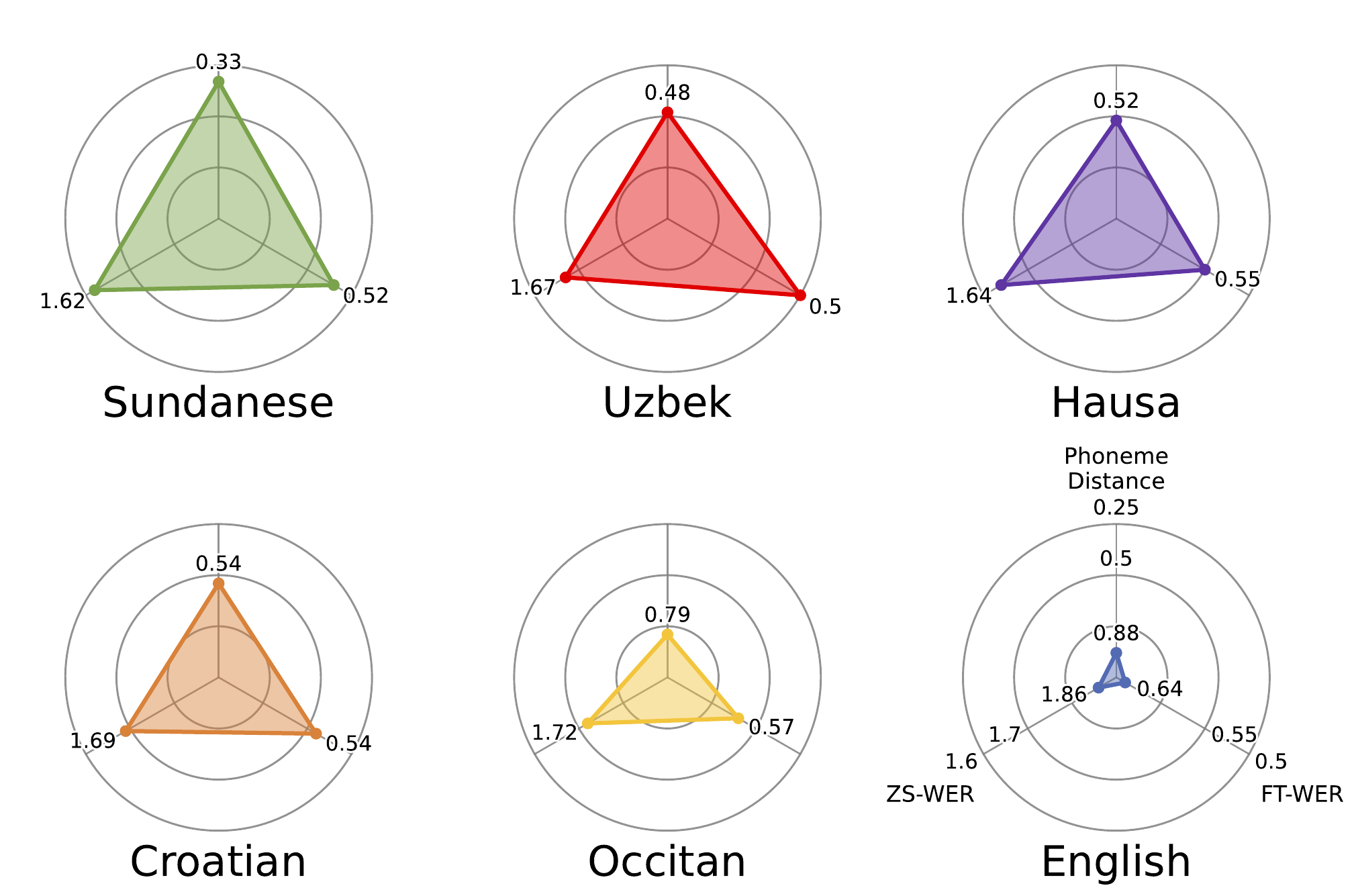}
  \caption{Similarity visualization of six candidate languages to Wardaman. To measure the segmental distance (\textbf{upward}) at the phonemic level between Wardaman and the six proxy languages, we compute the Hamming distance between their segmental inventories extracted from PHOIBLE \cite{phoible}. We also show the translation word error rate (WER) for these six proxy languages in both zero-shot (\textbf{leftward}) and fine-tuning (\textbf{rightward}) scenarios as a reference. A smaller phoneme distance indicates higher similarity, and a lower WER indicates stronger translation performance. 
  We observe that as the phoneme distance decreases, the performance of the translation model overall improves, verifying our model initialization design. 
  }
  \label{fig:proxy_phonological_similarity}
  \vskip -3mm
\end{figure}
\section{Method}
As shown in Fig.~\ref{fig:pipeline}, the proposed WARDEN system is composed of two separate stages: a transcription stage and a translation stage. Stage 1 turns a Wardaman audio into phonetic transcript, while Stage 2 translates the transcript into English. In this section, we detail the design of these two stages.

\subsection{Transcription Stage}
\label{subsec:asr}
In the first stage, to convert Wardaman speech audio into phonetic transcriptions, we fine-tune the Whisper-large-v3 \cite{radford2023whisper} model. Since Wardaman is not included in the pre-training data, it would be difficult to directly fine-tune Whisper on extremely limited data (6 hours in our setting).
To address this, we select a \textbf{phonetically similar proxy language} from Whisper's supported languages, which could help to speed up the knowledge transfer from phonetically similar languages \cite{conneau2020unsupervised}. 

To identify the most similar proxy language, we use the phonological system in the PHOIBLE database \cite{mor14phoible}, which encodes phoneme inventories as binary vectors. We then compute similarities between Wardaman and Whisper-supported languages using Hamming distance.
We identify the the following languages as closest phonetic matches to Wardaman: Sundanese, Uzbek, Hausa, Croatian, Occitan and English.

As shown in Figure~\ref{fig:proxy_phonological_similarity}, among all candidate languages, Sundanese has the smallest phoneme distance to Wardaman, indicating that their phonological structures are most similar. Furthermore, in zero-shot settings (no training), initialization with Sundanese as the proxy yields the loweset Word Error Rate (WER). When fine-tuned, this setting also demonstrates the second lowest in WER. Therefore, we reuse the Sundanese tag (\texttt{<su>})  for Wardaman language token initialization.

\begin{figure}[t]
  \centering
  \includegraphics[width=\linewidth]{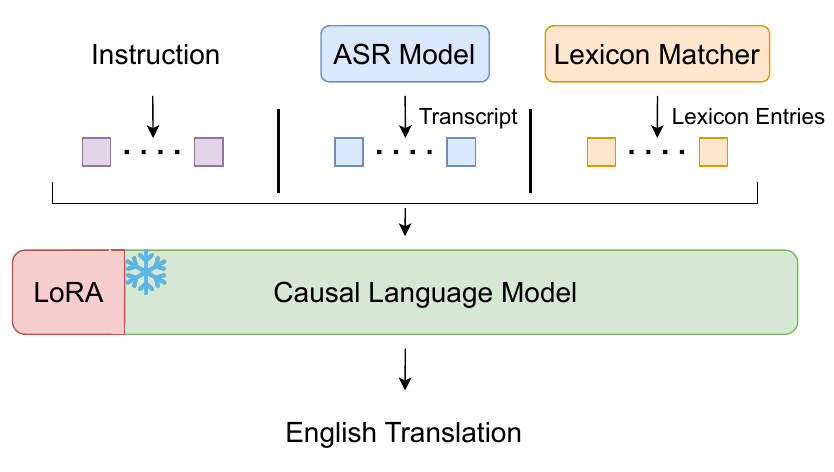}
  \vskip -3mm 
  \caption{LLM input organization for lexicon-augmented translation. The prompt combines a system instruction, the ASR transcript, and matched lexicon entries. The LLM is fine-tuned with low rank adaptation (LoRA) \cite{hu2022lora} to output English translations conditioned on this enriched context.}
  \label{fig:qwen_input}
  \vskip -2mm 
\end{figure}

\begin{figure*}[t]
  \centering
  \includegraphics[width=0.8\linewidth]{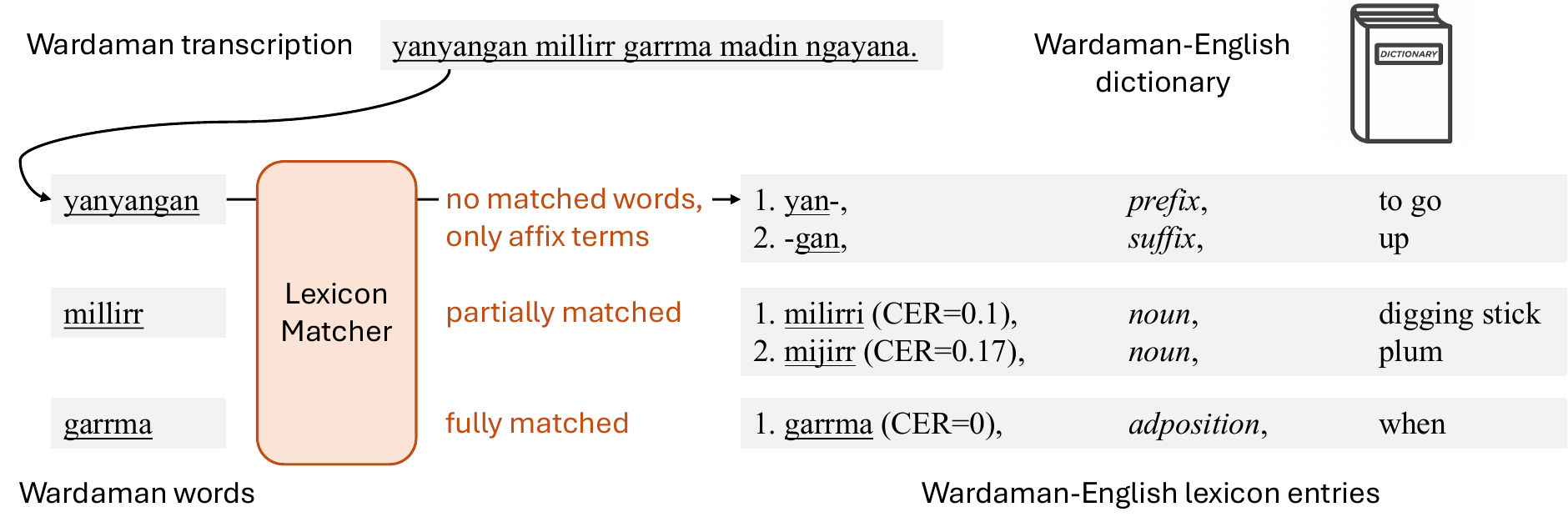}
  \caption{An example of lexicon matching. For each word in the Wardaman transcription result, the matcher retrieves the most relevant lexicon entries using CER and affix matching. 
  The resulting lexical cues are formatted and fed into the LLM to guide translation.}
  \label{fig:lexicon_matcher}
\end{figure*}

\subsection{Translation Stage}
\label{subsec: Translation}
This stage converts phonetic transcription to English sentences. 
Due to the lack of data, directly fine-tuning data-hungry translation models like Whisper \cite{liu2024whisper} and LLMs \cite{yang2025qwen3} yields poor results. 

To solve this issue, we propose to use a Wardaman-English dictionary compiled by linguists, used together with the transcript as input to the translation LLM. 
Our lexicon-enhanced workflow is shown in Figure~\ref{fig:qwen_input}. 
Given the transcription output and the Wardaman-English dictionary, for each word in the transcription, we use a lexicon matcher to retrieve the most relevant top-$k$ lexicons based on character error rate (CER). We then use them as input to the LLM translation model. This procedure gives preliminary translation results on the word level to the LLM. It thus helps ease the burden on LLM and alleviate overfitting. 
Details are provided as follows. 


\subsubsection{Wardaman-English Dictionary}

We first clean a Wardaman-English dictionary with approximately 2300 entries from FLEx \cite{silFieldWorksLanguage}. For each recorded Wardaman word, we include the following: part-of-speech tags, variants, definitions, and sentence examples. For the speech tag, we also include bound morphemes (\textit{e.g.}, prefix \texttt{ya-}, suffix \texttt{-yi}) as affix terms. This prepares the dictionary for LLMs with Wardaman words as \textit{keys} and their details and meanings as \textit{values}.

\subsubsection{Lexicon Matcher}
\label{subsubsec: Lexicon Matcher}
Since there is currently no semantic embedding model like BERT \cite{devlin2019bert} suitable for Wardaman, it is infeasible to match entries based on semantic similarity. To address this issue, as shown in Figure~\ref{fig:lexicon_matcher}, 
 we design a rule-based lexicon matcher that associates ASR transcription outputs with the most relevant lexicon entries in the following ways:

\begin{itemize}
\item \textbf{CER-based matching}: For each word in the transcription output, we compute its CER to all entries in the dictionary; the top-$k$ candidate words with CER $< \tau$ are retrieved from the lexicon.

\item \textbf{Affix matching}: Prefix/suffix matching is performed positionally (e.g., words starting with prefix \texttt{ya-}), disregarding CER to cover derived forms in transcription.

\end{itemize}

\subsubsection{LLM-Based Translation}
\label{subsubsec: LLM-Based Translation}

The matched entries are formatted as: \texttt{word (CER), part of speech, gloss}. The \texttt{word (CER)} term reports the retrieved word from the dictionary and its CER to the transcription output. The \texttt{gloss} term includes the definition of the Wardaman word. Both the ASR transcription output and the matched dictionary items are used as input to the LLM translation model. 

We use the system prompt to explain the task and the input format, which is listed below:

\begin{quote}
\textit{“Please translate the following Wardaman sentence into English, using the provided lexicons. Each lexicon entry is given in the form of word (character error rate), part of speech, and gloss.”}
\end{quote}

The exact transcription input and its corresponding lexicon terms are fed in to the user prompt listed as follows:

\begin{quote}
\textit{“Transcription: \{transcription\}. Lexicon entries: \{lexicon entries\}.”}
\end{quote}


We fine-tune the LLM using LoRA~\cite{hu2022lora}, enabling the model to learn how to generate content based on lexical cues. This transforms the LLM into a knowledge-based translator, capable of accurate translation even when the ASR output contains errors or word forms outside the vocabulary. 

Besides, we use data augmentation for translation training. 
First, we consider two types of input audio segmentation:
one based on short, naturally segmented utterances, and another formed by concatenating consecutive utterances into longer sequences to simulate extended spoken passages.
Second, we adopt the Whisper transcription output as a type of noisy data during the translation stage. This offers the translation system a glimpse of the transcription errors it might encounter.
We combine these two augmentation strategies during training, and report 
translation performance on long-format Whispers outputs by default.


\section{Results and discussion}
\subsection{Dataset}

\begin{table}[t]
\caption{Statistics of the processed Wardaman corpus.}
\label{tab:dataset_stats}
\setlength{\tabcolsep}{6pt}
\centering
\begin{tabular}{lccccc}
\toprule
{Category} & {Metric} & {Total} & {Avg.} & {Range} \\
\midrule
audio & duration (s) & 23,436 & 24.87 & 5.9--28.0 \\
transcription & \#words  & 30,490 & 31.93 & 2--91 \\
translation & \#words  & 29,966 & 49.94 & 3--99 \\
\bottomrule
\end{tabular}
\vspace{-0.5em}
\end{table}

Data in this paper comes from a long-term anthropological linguistic documentation project on the Wardaman language from 1976 to 2025. The corpus consists of audiovisual recordings that document biographical, mythological, historical narratives, and place-linked songs in this disappearing language. 

We construct a multimodal corpus of the Wardaman language based on recordings from Francesca Merlan’s fieldwork, which form the basis of her authoritative work on Wardaman grammar~\cite{merlan2011grammar}. The original material comprises 420 audio/video files, 
recorded over decades in indoor and outdoor environments in the Katherine region of the Northern Territory using handheld audio and video recorders. We only consider the subset accompanied by time-aligned ELAN annotation files (\texttt{.eaf}), providing linguistically verified Wardaman transcriptions and English translations. This enables fine-grained speech–text alignment critical for ASR modeling.


\begin{table}[t]
\centering
\caption{Comparing the transcription performance between WARDEN (last row) with various other models and training strategies. }
\label{tab:transcription_baselines}
\small
\setlength{\tabcolsep}{6pt}
\begin{tabular}{l|c|c|c}
\toprule
{Model} & {Fine-tuned} & {Special Init.} & {WER} $\downarrow$\\
\midrule
Speech2Text & \ding{55} & \ding{55} & 2.16 \\
Wav2Vec2 & \ding{55} & \ding{55} & 1.93 \\
Wav2Vec2 & \ding{52} & \ding{55} & 0.81 \\
Whisper & \ding{55} & \ding{55} & 1.62 \\
Whisper & \ding{52} & \ding{55} & 0.64 \\
\textbf{Whisper (ours)} & \ding{52} & \ding{52} & \textbf{0.52} \\
\bottomrule
\end{tabular}
\end{table}

\begin{figure}[t]
  \centering
  \includegraphics[width=\linewidth]{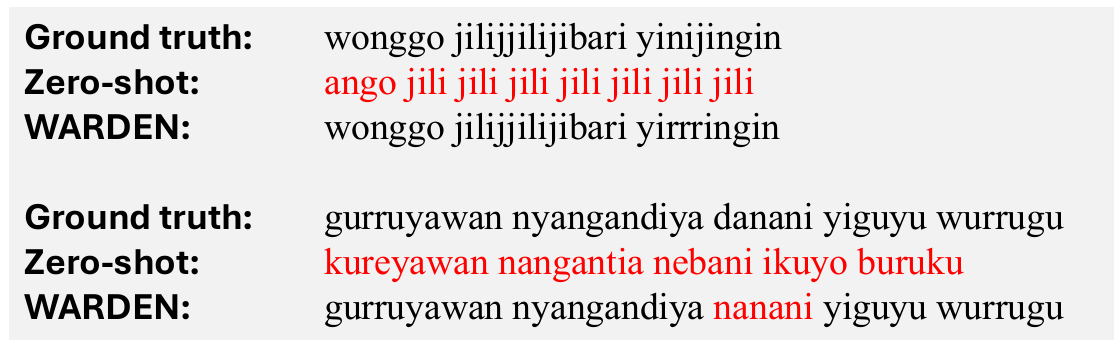}
  \caption{Qualitative comparison of Wardaman speech transcription outputs. Words highlighted in red indicate transcription errors (substitutions, insertions, or deletions) that increase the WER.}
  \label{fig:transcription_qual}
\end{figure}

Each ELAN file contains a hierarchical annotation structure; we extract only the primary transcription and translation tiers, discarding meta-annotations. Since Whisper accepts inputs up to 30 seconds, we concatenate adjacent ELAN segments within the same source file until approaching this limit-preserving linguistic coherence while satisfying model constraints. Critically, concatenation never crosses file boundaries, eliminating acoustic or speaker overlap between training and test sets and preventing data leakage. After filtering unusable sessions, the final dataset contains 98 source recordings, yielding 956 training samples with a total duration of approximately 6 hours (23,436 seconds). Table~\ref{tab:dataset_stats} shows the specific statistics after processing. Additionally, we export and manually clean the Wardaman–English lexicon from FLEx, containing approximately 2,000 entries with part-of-speech tags and glosses, covering about 30\% of the vocabulary in the corpus.


\subsection{Implementation Details}
For transcription, we fine-tune the first-stage Whisper-large-v3 model on eight 3090 GPUs using a full-parameter approach, where all the Whisper model parameters are fully fine-tuned. To ensure sufficient GPU memory, we use deepspeed zero-2 training with a learning rate of 0.0001 and a batch size of 4. 

For translation, we set the top-$k$ selection in the lexicon matcher to $k=3$, and the CER threshold $\tau=0.2$. For Qwen3 models, we use LoRA fine-tuning. We also use deepspeed zero-2 training with a learning rate of 0.001 and a batch size of 2.

\subsection{Experimental Results}
\label{subsec:transcription_finetuned}


\textbf{Transcription performance comparison.}
\label{subsubsec:transcription_vs_others}
In WARDEN, we fine-tune Whisper with the Sundanese token initialization.
In this section, we compare the transcription performance of WARDEN with other pretrained or fine-tuned models. Results are shown in Table~\ref{tab:transcription_baselines}. We have two major observations. 

\textbf{First}, WARDEN has the best transcription performance than the ordinary  Whisper fine-tuning which ranks second, where our WER is 0.12 lower. This demonstrates the effectiveness of using Sundanese token initialization. \textbf{Second}, fine-tuning on the Wardaman language improves zero-short performance significantly, on both the Wav2Vec2 model and Whisper model. This is expected because Wardaman is not used for pre-training the models. 


For qualitative results, as shown in Figure~\ref{fig:transcription_qual}, WARDEN yields transcription results with much fewer errors. For the incorrectly predicted transcriptions, they sound similar to the ground truth \textit{e.g.}, `buruku' versus `wurrugu'. 

\textbf{Translation performance comparison.}
\label{subsec:translation_vs_sota}
We compare the quantitative translation performance of WARDEN with competing methods in Table ~\ref{tab:baseline_sota_models}. In `few-shot', we use three Wardaman-English sentence pairs as examples for the LLM to do in-context learning, and no fine-tuning or lexicon matching is used. The oracle method is the last row where we directly use the ground-truth transcription results as LLM input. We have the following findings.

\begin{table}[t]
\centering
\caption{Translation performance comparison under different models and strategies. `Audio' indicates whether audio is directly used as model input. `GT' indicates whether the ground-truth transcription is used during translation inference. `Lexicon' means the use of lexicon matching in Section \ref{subsec: Translation}. `Few-shot' means directly using Wardaman-English sentence pairs from the dictionary for LLM in-context learning without fine-tuning LLMs.}

\label{tab:baseline_sota_models}
\setlength{\tabcolsep}{2.5pt}
\scriptsize
\begin{tabular}{l|c|c|c|c|c|c}
\toprule
{Model} & {Input} & {GT} & {Fine-tuned} & {Few-shot} & {Lexicon} & {BLEU-4} $\uparrow$\\
\midrule
Whisper & audio & - & \ding{52} & \ding{55} & \ding{55} & 1.42 \\
Qwen3-235B & text & \ding{55} & \ding{55} & \ding{55} & \ding{52} & 5.91 \\
Qwen3-235B & text & \ding{55} & \ding{55} & \ding{52} & \ding{55} & 6.34 \\
GPT-5 & text & \ding{55} & \ding{55} & \ding{55} & \ding{52} & 7.54 \\
GPT-5 & text & \ding{55} & \ding{55} & \ding{52} & \ding{55} & 7.19 \\
Qwen3-8B & text & \ding{55} & \ding{52} & \ding{55} & \ding{55} & 6.12 \\
Qwen3-8B & text & \ding{55} & \ding{55} & \ding{52} & \ding{55} & 3.77 \\
\textbf{Qwen3-8B (ours)} & text & \ding{55} & \ding{52} & \ding{55} & \ding{52} & \textbf{12.40} \\ \hline
\textcolor{gray}{Qwen3-8B (oracle)} & \textcolor{gray}{text} & \textcolor{gray}{\ding{52}} & \textcolor{gray}{\ding{52}} & \textcolor{gray}{\ding{55}} & \textcolor{gray}{\ding{52}} & \textcolor{gray}{16.42} \\
\bottomrule
\end{tabular}
\end{table}

\begin{figure}[t]
  \centering
  \includegraphics[width=\linewidth]{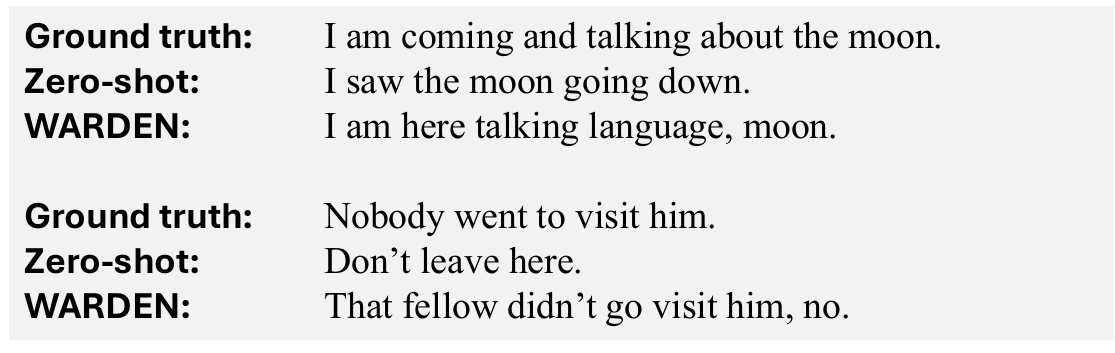}
  \caption{Qualitative comparison of translation outputs. }
  \label{fig:translation_qual}
\end{figure}

\textbf{First}, using lexicon matching results as additional LLM input improves normal LLM fine-tuning. Comparing Qwen3-8B (ours) and Qwen3-8B fine-tuning, the BLEU score improvement is +6.28. This demonstrates the effectiveness of injecting dictionary into the low-resource translation system. 

\textbf{Second}, when we use lexicon conditioning and fine-tune LLMs with translation data, we report the best translation performance BLEU = 12.40 using Qwen3-8B. The BLEU score is +4.86 higher than the best GPT-5 method. 

\textbf{Third}, fine-tuning Whisper using audio as input is the least effective. The BLEU score is only 1.42, much lower than fine-tuning LLMs for translation. 

\textbf{Last}, when no fine-tuning is performed, lexicon entry conditioning has similar performance with few-shot conditioning. For example, on Qwen3-235B, few-shot in-context learning is superior, while on GPT-5, lexicon conditioning without fine-tuning is better. Here, we note that lexicon conditioning uses much fewer tokens (61\% reduction in 3-shot settings) because the matched entries are much shorter than full sentences.

We show qualitative comparisons in Figure~\ref{fig:translation_qual}. We find that translations provided by WARDEN are more semantically aligned with the ground truth. Although the LLM could anchor the semantics of some words through lexicon in zero-shot strategy, a fine-tuned model better understands the whole sentence and more accurately connects the semantics of words matched by lexicon.

\begin{table}[t]
\centering
\caption{Ablation study on lexicon conditioning and fine-tuning. 
}
\label{tab:lexicon_effect_summary}
\setlength{\tabcolsep}{6pt}
\begin{tabular}{l|c|c|c}
\toprule
{Model} & {Fine-tuned} & {Lexicon} & {BLEU-4} \\
\midrule
Qwen3-8B & \ding{55} & \ding{55} & 1.97 \\
Qwen3-8B & \ding{55} & \ding{52} & 2.83 \\
Qwen3-8B & \ding{52} & \ding{55} & 6.12 \\
Qwen3-8B & \ding{52} & \ding{52} & \textbf{12.40} \\
\bottomrule
\end{tabular}
\end{table}

\begin{table}[t]
\centering
\caption{Ablation study on training augmentations.}
\label{tab:input_variant_ablation}
\setlength{\tabcolsep}{9pt}
\begin{tabular}{l|c|c|c}
\toprule
{Model} & {Short} & {Pred.} & {BLEU-4} \\
\midrule
Qwen3-8B & \ding{55} & \ding{55} & 6.17 \\
Qwen3-8B & \ding{52} & \ding{55} & 10.21 \\
Qwen3-8B & \ding{55} & \ding{52} & 11.96 \\
Qwen3-8B & \ding{52} & \ding{52} & \textbf{12.40} \\
\bottomrule
\end{tabular}
\end{table}

\begin{table}[t]
\centering
\caption{Variant study on lexicon injection strategy. Rows specify CER thresholds and columns specify $k$ in top-k selection. BLEU-4 is reported. }
\label{tab:lexicon_selection_grid}
\setlength{\tabcolsep}{8.2pt}
\begin{tabular}{l|ccccc}
\toprule
{CER} & {Top-1} & {Top-2} & {Top-3} & {Top-4} & {Top-5} \\
\midrule
0.1 & 9.85 & 10.60 & 10.72 & 10.89 & 10.93 \\
0.2 & 10.24 & 11.76 & \textbf{12.40} & 10.97 & 10.40 \\
0.3 & 9.63 & 10.94 & 11.82 & 10.49 & 9.85 \\
0.4 & 8.40 & 9.34 & 8.95 & 8.11 & 7.62 \\
0.5 & 8.26 & 8.50 & 8.21 & 7.30 & 7.07 \\
\bottomrule
\end{tabular}
\end{table}

\subsection{Ablation and Variant Studies}
\label{sec:translation_results}

We conduct ablation and variant studies to assess the contribution of key components in our pipeline, including cross-lingual initialization, lexicon integration, input augmentation, and lexicon selection strategies.

\textbf{Effect of Sundanese initialization.}
\label{subsubsec:ablation_sundanese}
For transcription, as shown in Table~\ref{tab:transcription_baselines}, removing Sundanese initialization leads to a WER increase of 0.12, verifying its effectiveness under the low-data regime. Moreover, our variant experiments (Fig.~\ref{fig:proxy_phonological_similarity}) show that Sundanese has the lowest phoneme distance, zero-shot WER and fine-tuned WER among all the candidate languages, further validating our choice. 


\textbf{Impact of lexicon integration.}
\label{subsubsec:ablation_lexicon}
As shown in Table 4, removing either the lexicon or fine-tuning reduces the BLEU score by 6.28 and 9.57, respectively. The score drops by 10.43 when both are missing. This confirms the effectiveness of lexicon in improving translation performance, especially with fine-tuning.


\textbf{Effect of input augmentation on translation.}
\label{subsubsec:ablation_augmentation}
Table \ref{tab:input_variant_ablation} shows that removing either short transcription samples or the samples predicted by the ASR model leads to a decrease in BLEU score by 0.44 and 2.19, respectively. Removing both results in a decrease of 6.23. This confirms the effectiveness of our data augmentation strategy for the translation task.


\textbf{Lexicon selection strategy for translation.}
The results from Table ~\ref{tab:lexicon_selection_grid} show that the translation performance on our data is best when the CER threshold is 0.2 and the top-3 candidates are selected.

\section{Conclusion and Future Work}

\label{sec:conclusion}

We propose WARDEN, a practical two-stage framework for transcribing and translating endangered languages using low-resource labeled data. Our system leverages a pre-trained Whisper model and uses Sundanese initialization before fine-tuning, because Sundanese is found to have a phonetic property similar to Wardaman. We show this method effectively decreases the word error rate in transcription. Then, we introduce a novel lexicon-enhanced translation stage, where we design a matching module to collect lexical knowledge related to transcription and fine-tune a large language model using the retrieved lexical entries as additional input. WARDEN can be effectively trained using only 6 hours of Wardaman audio, outperforming larger-scale zero-shot or fine-tuned models.

By improving the accuracy and accessibility of speech recognition tools for low-resource and endangered languages, we hope to support field linguists in the efficient transcription and translation of audio-visual recordings. Any potential contribution to community-led language documentation and revitalisation would depend on a range of factors, including the target translation language, the available materials, and the needs and preferences of the speech community. We therefore welcome input and feedback from Indigenous communities.

\textbf{Acknowledgement.} We thank Professor Francesca Merlan for her decades of work on the Wardaman language and for her advice and contributions to this project at various stages. We are also grateful for her sharing of the public-access Wardaman collection in ELAR and for her feedback on linguistic matters. We acknowledge the Wardaman people and community, and pay our respects to Elders past and present. We recognise that the language data belong to the speakers and their community. Any remaining errors are our own.


\bibliographystyle{IEEEtran}
\bibliography{mybib}

@article{liu2024whisper,
  title={Exploration of Whisper fine-tuning strategies for low-resource ASR},
  author={Liu, Yunpeng and Yang, Xukui and Qu, Dan},
  journal={EURASIP Journal on Audio, Speech, and Music Processing},
  volume={2024},
  number={1},
  pages={29},
  year={2024},
  publisher={Springer}
}

@article{sicard2023swissgerman,
  title={Fine-tuning whisper on low-resource languages for real-world applications},
  author={Timmel, Vincenzo and Paonessa, Claudio and Kakooee, Reza and Vogel, Manfred and Perruchoud, Daniel},
  journal={arXiv preprint arXiv:2412.15726},
  year={2024}
}

@online{merlan_wardaman_elar,
  author       = {Merlan, Francesca},
  title        = {Wardaman Dictionary, Narrative, Song and Country},
  year         = {2025},
  organization = {Endangered Languages Archive (ELAR)},
  url          = {http://hdl.handle.net/2196/884f9353-ea4c-4686-b83c-18cdb828193z},
  urldate      = {2026-03-15}
}

@book{merlan1994narratives,
url = {https://doi.org/10.1515/9783110871371},
title = {A Grammar of Wardaman},
title = {A Language of the Northern Territory of Australia},
author = {Francesca C. Merlan},
publisher = {De Gruyter Mouton},
address = {Berlin, New York},
doi = {doi:10.1515/9783110871371},
isbn = {9783110871371},
year = {1994},
lastchecked = {2026-05-07}
}

@book{phoible,
  address   = {Jena},
  editor    = {Steven Moran and Daniel McCloy},
  publisher = {Max Planck Institute for the Science of Human History},
  title     = {PHOIBLE 2.0},
  url       = {https://phoible.org/},
  year      = {2019}
}

@article{yang2025qwen3,
  title={Qwen3 technical report},
  author={Yang, An and Li, Anfeng and Yang, Baosong and Zhang, Beichen and Hui, Binyuan and Zheng, Bo and Yu, Bowen and Gao, Chang and Huang, Chengen and Lv, Chenxu and others},
  journal={arXiv preprint arXiv:2505.09388},
  year={2025}
}

@article{yong2024lexcgen,
  title={Lexc-gen: Generating data for extremely low-resource languages with large language models and bilingual lexicons},
  author={Yong, Zheng-Xin and Menghini, Cristina and Bach, Stephen H},
  journal={arXiv preprint arXiv:2402.14086},
  year={2024}
}

@inproceedings{zheng2025tangut,
  title={Incorporating Lexicon-Aligned Prompting in Large Language Model for Tangut--Chinese Translation},
  author={Zheng, Yuxi and Yu, Jingsong},
  booktitle={Proceedings of the Second Workshop on Ancient Language Processing},
  pages={127--136},
  year={2025}
}

@article{hu2022lora,
  title={Lora: Low-rank adaptation of large language models.},
  author={Hu, Edward J and Shen, Yelong and Wallis, Phillip and Allen-Zhu, Zeyuan and Li, Yuanzhi and Wang, Shean and Wang, Lu and Chen, Weizhu and others},
  journal={ICLR},
  volume={1},
  number={2},
  pages={3},
  year={2022}
}

@inproceedings{radford2023whisper,
  title={Robust speech recognition via large-scale weak supervision},
  author={Radford, Alec and Kim, Jong Wook and Xu, Tao and Brockman, Greg and McLeavey, Christine and Sutskever, Ilya},
  booktitle={International conference on machine learning},
  pages={28492--28518},
  year={2023},
  organization={PMLR}
}

@book{mor14phoible,
  address   = {Jena},
  editor    = {Steven Moran and Daniel McCloy},
  publisher = {Max Planck Institute for the Science of Human History},
  title     = {PHOIBLE 2.0},
  url       = {https://phoible.org/},
  year      = {2019}
}

@inproceedings{devlin2019bert,
  title={Bert: Pre-training of deep bidirectional transformers for language understanding},
  author={Devlin, Jacob and Chang, Ming-Wei and Lee, Kenton and Toutanova, Kristina},
  booktitle={Proceedings of the 2019 conference of the North American chapter of the association for computational linguistics: human language technologies, volume 1 (long and short papers)},
  pages={4171--4186},
  year={2019}
}

@incollection{himmelmann2006language,
  title     = {Language Documentation: What is it and what is it good for?},
  author    = {Himmelmann, Nikolaus P.},
  booktitle = {Essentials of Language Documentation},
  editor    = {Gippert, Jost and Himmelmann, Nikolaus P. and Mosel, Ulrike},
  pages     = {1--30},
  year      = {2006},
  publisher = {Mouton de Gruyter},
  address   = {Berlin}
}

@inproceedings{adams2019massively,
  title     = {Massively Multilingual Speech Recognition for Endangered Languages},
  author    = {Adams, Oliver and Kjellstr{\"o}m, Hedvig and others},
  booktitle = {Proceedings of Interspeech},
  pages     = {2050--2054},
  year      = {2019},
  doi       = {10.23919/INTERSPEECH.2019.8932730}
}

@inproceedings{bird2022machine,
  title     = {Machine Translation for Indigenous Languages: Challenges and Opportunities},
  author    = {Bird, Steven and Hanke, Florian and Adams, Oliver and others},
  booktitle = {Proceedings of the Workshop on Language Technologies for Indigenous Languages (LT4IL)},
  pages     = {1--10},
  year      = {2022},
  url       = {https://aclanthology.org/2022.lt4il-1.1}
}

@inproceedings{conneau2020unsupervised,
  title     = {Unsupervised Cross-lingual Representation Learning for Speech Recognition},
  author    = {Conneau, Alexis and Baevski, Alexei and Collobert, Ronan and Mohamed, Abdelrahman and Auli, Michael},
  booktitle = {Proceedings of Interspeech},
  pages     = {3166--3170},
  year      = {2020},
  doi       = {10.23919/INTERSPEECH.2020.84}
}

@misc{silFieldWorksLanguage,
	author = {},
	title = {{F}ield{W}orks {L}anguage {E}xplorer™ - {D}ictionary {C}reation {S}oftware --- software.sil.org},
	howpublished = {\url{https://software.sil.org/fieldworks/}},
	year = {},
	note = {[Accessed 02-04-2026]},
}

@book{merlan2011grammar,
  title={A grammar of Wardaman: A language of the Northern Territory of Australia},
  author={Merlan, Francesca C},
  volume={11},
  year={2011},
  publisher={Walter de Gruyter}
}

\end{document}